\documentclass[twocolumn]{autart}    

\usepackage{graphicx}          

\usepackage{amsmath}
\usepackage{amssymb}
\usepackage{mathtools}

\usepackage[noadjust]{cite}

\newtheorem{theorem}{Theorem}
\newtheorem{lemma}{Lemma}
\newtheorem{proposition}{Proposition}
\newtheorem{corollary}{Corollary}

\theoremstyle{definition}
\newtheorem{definition}{Definition}
\newtheorem{assumption}{Assumption}
\newtheorem{remark}{Remark}
\newtheorem{problem}{Problem}

\begin{document}

\begin{frontmatter}

\title{Safe Exploration in Model-based Reinforcement Learning using Control Barrier Functions \thanksref{footnoteinfo}} 

\thanks[footnoteinfo]{This paper was not presented at any IFAC 
meeting. Corresponding author M.~H.~Cohen.}

\author[BU]{Max H. Cohen}\ead{maxcohen@bu.edu},
\author[BU]{Calin Belta}\ead{cbelta@bu.edu}
\address[BU]{Department of Mechanical Engineering, Boston University, 110 Cummington Mall, Boston, MA, USA}
          

\begin{abstract}                          
    This paper develops a model-based reinforcement learning (MBRL) framework for learning online the value function of an infinite-horizon optimal control problem while obeying safety constraints expressed as control barrier functions (CBFs). Our approach is facilitated by the development of a novel class of CBFs, termed Lyapunov-like CBFs (LCBFs), that retain the beneficial properties of CBFs for developing minimally-invasive safe control policies while also possessing desirable Lyapunov-like qualities such as positive semi-definiteness. We show how these LCBFs can be used to augment a learning-based control policy to guarantee safety and then leverage this approach to develop a safe exploration framework in a MBRL setting. We demonstrate that our approach can handle more general safety constraints than comparative methods via numerical examples. 
\end{abstract}

\end{frontmatter}

\section{Introduction}
Learning-based control methods, such as reinforcement learning (RL), have shown success in solving complex control problems. Although such methods have demonstrated success in simulated environments, the lack of provable safety guarantees has limited the application of these techniques to real-world safety-critical systems. To address these challenges, the field of \emph{safe learning-based control} has emerged with the objective of ensuring the safety of learning-enabled systems. The concept of safety in control systems is often formalized through the forward invariance \cite{BlanchiniAutomatica99} of designated safe sets. Popular approaches to safe learning include reachability analysis \cite{FisacTAC19}, model predictive control \cite{ZeilingerAR20,GrosTAC20,ChuACC18}, and barrier functions (BFs) \cite{MurrayAAAI19,TaylorL4DC20}. Similar to how Lyapunov functions are used to study the stability of equilibrium points without computing a system's solution, BFs \cite{AmesTAC17,AmesECC19,AllgowerSNCS07,PanagouTAC16,TeeAutomatica09,WillisAutomatica04} allow for one to study the invariance of sets without explicitly computing a system's reachable set. Motivated by the need for general tools that facilitate the design of controllers with safety guarantees, the concept of a control BF (CBF) was introduced in \cite{AllgowerSNCS07,AmesTAC17,AmesECC19}, and has been successfully applied to systems such as autonomous vehicles and multi-agent systems \cite{AmesECC19}.

In this paper we unite CBFs and model-based reinforcement learning (MBRL) to develop a safe exploration framework for jointly learning online the dynamics of an uncertain control affine system and the optimal value function/policy of an infinite-horizon optimal stabilization problem. Our approach is facilitated by the class of Lyapunov-like barrier functions introduced in \cite{PanagouTAC16}, which are used to develop a partially model-free robust safeguarding controller that can be combined with an arbitrary learning-based control policy to guarantee safety. This safeguarding controller is leveraged to build upon the MBRL framework from \cite{KamalapurkarAutomatica16-mbrl,KamalapurkarAutomatica16-staf,Kamalapurkar} to develop a safe exploration scheme in which the value function is learned online via ``simulation of experience." In this approach, the tension between exploration and safety is addressed by simulating on-the-fly an approximated model of the system at unexplored points in the state space to generate data for learning the value function without risking safety violation of the original system. We provide proofs of convergence of this approximation scheme and demonstrate numerically the advantages of our approach over related safe online RL approaches.

\paragraph*{Related work}
The online RL method considered herein is rooted in the seminal work of \cite{VamvoudakisAutomatica10}, where techniques from adaptive control \cite{Krstic} were used to solve online an unconstrained infinite-horizon optimal control problem for a nonlinear system with known dynamics. This approach was quickly extended to uncertain systems using model-free and model-based RL methods (see \cite{LewisCSS12,LewisTNNLS17} for surveys of model-free methods and \cite{Kamalapurkar} for a monograph of model-based methods). Although various extensions have been proposed over the past decade, these techniques have been limited to unconstrained problems or those with actuator constraints \cite{VamvoudakisTNNLS15,DeptulaAutomatica20}. More recently, these RL techniques have begun to consider \emph{safety constraints} by incorporating BF-based terms in the problem's cost function \cite{VamvoudakisIJRNC20,GreeneLCSS20,KamalapurkarACC21,CohenCDC20,KiumarsiIJRNC21}. The approaches from \cite{VamvoudakisIJRNC20,GreeneLCSS20,KamalapurkarACC21} rely on a barrier Lyapunov function (BLF) \cite{TeeAutomatica09} based system transformation to map a constrained optimal control problem to an unconstrained one, which can then be solved using existing approaches. However, these techniques are limited to rectangular state constraints (i.e., the safe set is a hyperrectangle), which do not encompass the complex safety specifications, such as collision avoidance, encountered in applications such as robotics. These BF-based techniques were generalized to safe sets defined by the zero-superlevel set of a general continuously differentiable function in \cite{CohenCDC20,KiumarsiIJRNC21} by including a CBF-based term in the problem's cost function, which was inspired by the motion planning framework from \cite{DeptulaTRO20}. However, as argued in \cite{KamalapurkarACC21}, the approaches from \cite{CohenCDC20,KiumarsiIJRNC21,DeptulaTRO20} are facilitated by the strong assumption that the resulting value function is continuously differentiable - a condition that may fail to hold for various systems and safe sets. Although the aforementioned approaches have demonstrated success in practice, an important limitation faced by \cite{VamvoudakisIJRNC20,GreeneLCSS20,KamalapurkarACC21,CohenCDC20,KiumarsiIJRNC21,DeptulaTRO20} in the context of safety-critical control is the use of the resulting value function, which acts as a BLF, as a safety certificate. Since the ultimate objective of these approaches is to learn the value function/safety certificate, safety and learning become tightly coupled - safety guarantees are conditioned upon convergence of the RL algorithm, which is predicated on excitation conditions that cannot be verified in practice. It is the aim of this paper to address this limitation by \emph{decoupling} learning from safety, with the latter being guaranteed at all times independent of any conditions associated with learning. Since the value function of the class of optimal control problems considered herein is a control Lyapunov function (CLF) \cite[Ch. 8.5]{Sontag}, our method can be seen as an approach to safely learn an optimal CLF for an uncertain system using data from a single trajectory. Compared to traditional CBF-based approaches that unite CLFs and CBFs to achieve dual objectives of stability and safety, this implies one need only to construct a valid CBF to achieve stability and safety, with the additional benefit of accounting for partially uncertain dynamics.

\paragraph*{Contributions}
The contributions of this paper are threefold. First, we introduce a new class of CBFs, termed Lyapunov-like CBFs (LCBFs), inspired by the Lyapunov-like barrier functions introduced in \cite{PanagouTAC16}, that retain the important properties of CBFs for making safety guarantees while possessing desirable Lyapunov-like qualities that facilitate the development of safe and stabilizing controllers. We illustrate how the gradient of this LCBF can be used to construct a safeguarding controller that shields a performance-driven RL policy in a minimally invasive fashion to guarantee safety. Second, we extend the MBRL architecture from \cite{KamalapurkarAutomatica16-staf} to develop a safe exploration framework in which the value function and optimal policy of an unconstrained optimal control problem are \emph{safely learned} online for an uncertain nonlinear system. {The fundamental distinction between our approach and those of related works \cite{VamvoudakisIJRNC20,GreeneLCSS20,KamalapurkarACC21,CohenCDC20,KiumarsiIJRNC21,DeptulaTRO20} is that the aforementioned methods aim to \emph{learn a safe policy}, whereas our safe exploration framework allows for \emph{safely learning a performance-driven policy} - an approach more aligned with works such as \cite{FisacTAC19,MurrayAAAI19}.} Although, at a high-level, our technical approach is similar to those of \cite{KamalapurkarAutomatica16-mbrl,KamalapurkarAutomatica16-staf}, the approach in this paper presents a departure from those in \cite{KamalapurkarAutomatica16-mbrl,KamalapurkarAutomatica16-staf} in that \emph{different} policies are used for exploration (learned policy) and deployment on the original system (safe policy), which complicates the resulting convergence analysis. Third, we present numerical examples demonstrating the improved safety guarantees of the proposed MBRL method compared to those that enforce safety through the minimization of a suitably constructed cost functional. In an additional numerical example, we illustrate empirically that our approach allows for simultaneous stabilization and obstacle avoidance via dynamic time-varying feedback.

\paragraph*{Notation}
A continuous function $\alpha\,:\,\mathbb{R}_{\geq0}\rightarrow\mathbb{R}_{\geq0}$ is said to be a class $\mathcal{K}$ function, denoted by $\alpha\in\mathcal{K}$, if it is strictly increasing and $\alpha(0)=0$. The derivative of a continuously differentiable function $f(x,y)$ with respect to its first argument is denoted by $\nabla f(x,y)=\tfrac{\partial f(x,y)}{\partial x}$ and is interpreted as a row vector. The operator $\|\cdot\|$ denotes the standard Euclidean norm, $I_n\in\mathbb{R}^{n\times n}$ denotes an $n\times n$ identity matrix, and $\lambda_{\max}(M),\,\lambda_{\min}(M)$ return the maximum and minimum eigenvalues of a matrix $M$, respectively. Given a signal $d\,:\,\mathbb{R}_{\geq0}\rightarrow\mathbb{R}^m$, we define $\|d\|_\infty\coloneqq\sup_t\|d(t)\|$. Given a set $\mathcal{C}$, the notation $\partial\mathcal{C}$ denotes the boundary of $\mathcal{C}$ and $\text{Int}(\mathcal{C})$ denotes the interior of $\mathcal{C}$. We define $\mathcal{B}_r(x_0)\coloneqq\{x\in\mathbb{R}^n\:|\:\|x-x_0\|\leq r \}$ as a closed ball of radius $r>0$ centered at $x_0\in\mathbb{R}^n$.

\section{Preliminaries and Problem Formulation}\label{sec:prelim}
Consider a nonlinear control affine system of the form
\begin{equation}\label{eq:dyn}
    \dot{x}=f(x)+g(x)u,
\end{equation}
where $x\in\mathbb{R}^n$ , $f\,:\,\mathbb{R}^n\rightarrow\mathbb{R}^n$ and $g\,:\,\mathbb{R}^n\rightarrow\mathbb{R}^{n\times m}$ are locally Lipschitz, and { $u\in\mathbb{R}^m$ is the control input. We assume that $f(0)=0$ so that the origin is in equilibrium point of the uncontrolled system.} Let $k(x,t)$ be a feedback control policy such that the closed-loop { vector field $f_{\text{cl}}(x,t)\coloneqq f(x)+g(x)k(x,t)$} is locally Lipschitz in $x$ and piecewise continuous in $t$. Under these assumptions, given an initial condition $x_0\coloneqq x(t_0)\in\mathbb{R}^n$ at time $t=t_0$, there exists some maximal interval of existence $\mathcal{I}(x_0)=[t_0,\tau_{\max})$ such that $x(t)$ satisfies
\begin{equation}\label{eq:closed_loop}
\dot{x}(t)=f(x(t))+g(x(t))k(x(t),t),\quad x(t_0)=x_0,
\end{equation}
for all $t\in\mathcal{I}(x_0)$. We say that a set $\mathcal{C}\subset\mathbb{R}^n$ is \emph{forward invariant} for \eqref{eq:closed_loop} if, for each $x_0\in\mathcal{C}$, the solution to \eqref{eq:closed_loop} satisfies $x(t)\in\mathcal{C}$ for all $t\in\mathcal{I}(x_0)$. If $\mathcal{C}$ is forward invariant for \eqref{eq:closed_loop}, we say the closed-loop system is \emph{safe} with respect to $\mathcal{C}$ and that $\mathcal{C}$ is a \emph{safe set}. In this paper, we consider candidate safe sets of the form 
\begin{subequations}\label{eq:safe_set}
\begin{align}
\mathcal{C}=&\{x\in\mathbb{R}^n\,|\,h(x)\geq0\},\\
\partial\mathcal{C}=&\{x\in\mathbb{R}^n\,|\,h(x)=0\},\\
\text{Int}(\mathcal{C})=&\{x\in\mathbb{R}^n\,|\,h(x)>0\},
\end{align}
\end{subequations}
where $h\,:\,\mathbb{R}^n\rightarrow\mathbb{R}$ is continuously differentiable. A powerful tool for designing control policies that render sets of the form \eqref{eq:safe_set} forward invariant is the concept of a \emph{control barrier function} (CBF) \cite{AmesTAC17,AmesECC19}, defined as follows:
\begin{definition}[\cite{AmesTAC17}]\label{def:CBF}
    A continuously differentiable function $b\,:\,\text{Int}(\mathcal{C})\rightarrow\mathbb{R}_{\geq0}$ is said to be a \emph{control barrier function} (CBF) for \eqref{eq:dyn} over a set $\mathcal{C}\subset\mathbb{R}^n$ as in \eqref{eq:safe_set} if there exist $\alpha_1,\,\alpha_2,\,\alpha_3\in\mathcal{K}$ such that for all $x\in\text{Int}(\mathcal{C})$
    \begin{subequations}
    \begin{equation}\label{eq:CBF1}
        \frac{1}{\alpha_1\left(h(x)\right)}\leq b(x) \leq \frac{1}{\alpha_2\left(h(x)\right)},
    \end{equation}
    \begin{equation}\label{eq:CBF2}
        { \inf_{u\in\mathbb{R}^m}}\{\nabla b(x)f(x) + \nabla b(x)g(x)u  \}\leq \alpha_3\left(h(x)\right).
    \end{equation}
    \end{subequations}
\end{definition}
We refer to a continuously differentiable function $b$ satisfying \eqref{eq:CBF1} as a \emph{candidate} CBF. The main result regarding CBFs is that the existence of such a function implies the existence of a policy $u=k(x)$ that renders $\text{Int}(\mathcal{C})$ forward invariant for \eqref{eq:dyn} \cite[Cor. 1]{AmesTAC17}.
\begin{problem}\label{prob:main}
    Consider system \eqref{eq:dyn} and a set $\mathcal{C}\subset\mathbb{R}^n$ as in \eqref{eq:safe_set}. Given a {candidate CBF $b\,:\,\text{Int}(\mathcal{C})\rightarrow\mathbb{R}_{\geq0}$}, find a control policy { $u=k(x)$} such that $\text{Int}(\mathcal{C})$ is forward invariant for \eqref{eq:dyn} and the origin is stable for \eqref{eq:dyn}.
\end{problem}
Our first step to addressing Problem \ref{prob:main} is to introduce a new class of CBFs that retain the crucial properties of CBFs but are also positive semi-definite on $\text{Int}(\mathcal{C})$, which becomes important when ensuring that the origin is an equilibrium point for the closed-loop system. To find a stabilizing control policy, we take a RL-based approach in which the value function of an optimal control problem is safely learned online.

\section{Lyapunov-like Control Barrier Functions}\label{sec:LCBF}
In this section, we propose a new class of CBFs and present various technical results that illustrate their properties. Based upon the development from \cite{PanagouTAC16}, we consider the following \emph{recentered} barrier function \cite{WillisAutomatica04}
\begin{equation}\label{eq:B}
    B(x)\coloneqq\left(b(x)-b(0) \right)^2,
\end{equation}
where $b\,:\,\text{Int}(\mathcal{C})\rightarrow\mathbb{R}_{\geq0}$ is a { candidate CBF with $0\in\text{Int}(\mathcal{C})$}. Throughout this paper {we refer to \eqref{eq:B} as a \emph{Lyapunov-like CBF} (LCBF)} { candidate} as $B$ can be shown to satisfy the important properties
\begin{equation*}
\begin{aligned}
    \inf_{x\in\text{Int}(\mathcal{C})} B(x)\geq0,  \qquad B(0)=0, \qquad  \lim_{x\rightarrow\partial\mathcal{C}}B(x)=\infty.
    \end{aligned}
\end{equation*}
The primary distinction between standard CBFs and LCBFs is that $B$ vanishes at the origin, which will become important when combining the subsequently designed safeguarding controller with a nominal stabilizing policy. The following lemma establishes conditions on $B$ that ensure the forward invariance of $\mathcal{C}$.

\begin{lemma}\label{lemma:B}
    Consider system \eqref{eq:closed_loop} and assume the origin is contained in $\text{Int}(\mathcal{C})$. Provided { $B(x(t))<\infty$ for all $t\in\mathcal{I}(x_0)$}, then $\text{Int}(\mathcal{C})$ is forward invariant for \eqref{eq:closed_loop}.
\end{lemma}
    
\begin{pf}
Since $b(x)$ shares the same properties as $1/\alpha(h(x))$ for $\alpha\in\mathcal{K}$ \cite[Remark 1]{AmesTAC17} and since $0\in\text{Int}(\mathcal{C})$ implies { $b(0)<\infty$, we have $B(x)<\infty\implies 1/\alpha(h(x))<\infty\implies h(x)>0$. Thus, if $B(x(t))<\infty,\,\forall t\in\mathcal{I}(x_0)$}, then $h(x(t))>0, \,\forall t\in\mathcal{I}(x_0)$, which implies $x(t)\in\text{Int}(\mathcal{C}),\,\forall t\in\mathcal{I}(x_0)$. \qed
\end{pf}

Similar to how CBFs are often used as a safety filter for a nominal control policy \cite{AmesECC19}, in this paper we exploit the properties of LCBFs to endow existing control policies for \eqref{eq:dyn} with strong safety guarantees. In particular, we propose the following \emph{safeguarding controller}
\begin{equation}\label{eq:kb}
    k_b(x)\coloneqq-c_bg(x)^T\nabla B(x)^T,
\end{equation}
where $c_b\in\mathbb{R}_{>0}$ is a gain and $B$ is a LCBF candidate. { The following assumption places conditions on the dynamics \eqref{eq:dyn}, the safe set $\mathcal{C}$, and candidate CBF $b$ that facilitate the development of subsequent results.

\begin{assumption}\label{assumption:dyn}
Given system \eqref{eq:dyn} and a candidate safe set $\mathcal{C}$, the following conditions hold:
\begin{enumerate} 
    \item There exists a positive, non-decreasing function $\ell_f\,:\,\mathbb{R}_{
    \geq0}\rightarrow\mathbb{R}_{\geq0}$ such that $\|f(x)\|\leq\ell_{f}(\|x\|)\|x\|$ for all $x\in\mathcal{C}$ and $\lim_{x\rightarrow\partial\mathcal{C}}\ell_{f}(\|x\|)\|x\|<\infty$.
    \item There exists a positive constant $\underline{g}\in\mathbb{R}_{>0}$ such that $\underline{g}\leq \|g(x)\|$ for all $x\in \mathcal{C}$.
    \item There exists a neighborhood of $\partial\mathcal{C}$, denoted by $\mathcal{N}(\partial\mathcal{C})$, such that $0\notin\mathcal{N}(\partial\mathcal{C})$ and $\|\nabla b(x)g(x)\|\neq 0$ for all $x\in\mathcal{N}(\partial\mathcal{C})$.
\end{enumerate}
\end{assumption}

\begin{remark}
The first condition in Assumption \ref{assumption:dyn}, made largely for technical reasons, places growth restrictions on the system drift and ensures the drift dynamics do not ``blow up" on the boundary of $\mathcal{C}$. The second assumption ensures the control directions do not vanish where control authority may be required to keep the system safe. The third condition is, in essence, a feasibility assumption that ensures the existence of control values that render the safe set forward invariant and is not restrictive provided the candidate CBF $b$ has relative degree one with respect to \eqref{eq:dyn}. As discussed later in Remark \ref{remark:CBF}, this last condition is stronger than necessary and can be slightly relaxed provided $b$ is a \emph{valid} CBF. Additionally, note that $\|\nabla b(x)g(x)\|\neq 0$ for all $x\in \mathcal{N}(\partial\mathcal{C})$ in conjunction with $0\notin \mathcal{N}(\partial\mathcal{C})$ implies $\|\nabla B(x)g(x)\|\neq0$ for all $x\in \mathcal{N}(\partial\mathcal{C})$.
\end{remark}
} The following theorem shows that, under Assumption \ref{assumption:dyn}, the policy in \eqref{eq:kb} renders $\text{Int}(\mathcal{C})$ forward invariant for \eqref{eq:dyn}.

\begin{theorem}\label{thm:LCBF}
Consider system \eqref{eq:dyn}, a set $\mathcal{C}\subset\mathbb{R}^n$ as in \eqref{eq:safe_set} with $0\in\text{Int}(\mathcal{C})$, and let $b\,:\,\text{Int}(\mathcal{C})\rightarrow\mathbb{R}_{\geq0}$ be a candidate CBF for \eqref{eq:dyn} on $\mathcal{C}$. Provided Assumption \ref{assumption:dyn} holds, the controller $u=k_b(x)$ with $k_b$ as in \eqref{eq:kb} renders $\text{Int}(\mathcal{C})$ forward invariant for the closed-loop system \eqref{eq:dyn}.
\end{theorem}

\begin{pf}
    Taking the derivative of $B$ along the closed-loop vector field yields
    \[
    \begin{aligned}
        \dot{B}(x)= &\nabla B(x)f(x) + \nabla B(x)g(x)k_b(x) \\
        = & \nabla B(x)f(x)- c_b\|\nabla B(x)g(x)\|^2.
    \end{aligned}
    \]
    Provided Assumption \ref{assumption:dyn} holds, $\dot{B}$ can be upper bounded for all $x\in \mathcal{N}(\partial\mathcal{C})$ as
    \begin{equation}\label{eq:Bdotbound}
    \begin{aligned}
    \dot{B}(x)\leq & \|\nabla B(x)\|\ell_{f}(\|x\|)\|x\| - c_b\|\nabla B(x) g(x)\|^2\\
    \leq & \|\nabla B(x)\|\ell_{f}(\|x\|)\|x\| -c_b\|\nabla B(x) \|^2\underline{g}^2 \\
    = & \|\nabla B(x)\|^2\cdot\left(\frac{\ell_{f}(\|x\|)\|x\|}{\|\nabla B(x)\|} - c_b\underline{g}^2  \right).
    \end{aligned}
    \end{equation}
    If $\ell_{f}(\|x\|)\|x\|$ is finite in the limit as $x\rightarrow \partial\mathcal{C}$ then ${\ell_{f}(\|x\|)\|x\|}/{\|\nabla B(x)\|}=0$ in the limit as $x\rightarrow \partial\mathcal{C}$, which implies that $({\ell_{f}(\|x\|)\|x\|}/{\|\nabla B(x)\|} - c_b\underline{g}^2)=- c_b\underline{g}^2$ in the limit as $x\rightarrow \partial\mathcal{C}$. Hence, taking limits in \eqref{eq:Bdotbound} as $x$ tends to $\partial\mathcal{C}$ yields
    \begin{equation}\label{eq:Bdotlim}
    \begin{aligned}
    \lim_{x\rightarrow\partial\mathcal{C}}\dot{B}(x)= -\infty < 0.
    \end{aligned}
    \end{equation}
    The fact that $\lim_{x\rightarrow\partial\mathcal{C}}\dot{B}(x)<0$ precludes the existence of trajectories that enter $\partial\mathcal{C}$. To see this, suppose there exists a trajectory of the closed-loop system $t\mapsto x(t)$ with $x_0\in\text{Int}(\mathcal{C})$ defined on some maximal interval of existence $\mathcal{I}(x_0)$ such that $\lim_{t\rightarrow T}B(x(t))=\infty$ for some finite $T\in\mathcal{I}(x_0)$. Since $B(x)\rightarrow \infty \iff b(x)\rightarrow\infty$, this implies that $\lim_{t\rightarrow T}x(t)\in\partial\mathcal{C}$. However, since $\lim_{t\rightarrow T}B(x(t))=\infty$ for some finite $T\in \mathcal{I}(x_0)$, this also implies that $\lim_{t\rightarrow T}\dot{B}(x(t))=\infty$, which contradicts \eqref{eq:Bdotlim}. Hence, there cannot exist a trajectory of the closed-loop system for which $\lim_{t\rightarrow T}B(x(t))=\infty$ for some finite $T\in\mathcal{I}(x_0)$, implying $B(x(t))<\infty$ for all $t\in\mathcal{I}(x_0)$. The forward invariance of $\text{Int}(\mathcal{C})$ follows from Lemma \ref{lemma:B}. \qed
\end{pf}
    
Having established the safety of \eqref{eq:dyn} under the influence of \eqref{eq:kb}, we show in the following corollary to Theorem \ref{thm:LCBF} how the safeguarding controller can be used to modify an existing control policy for \eqref{eq:dyn} to guarantee safety.

\begin{corollary}\label{cor:safe_control}
{ Let $k(x,t)$ be a nominal control policy, locally Lipschitz in $x$ and piecewise continuous in $t$, satisfying $k(0,t)=0$ for all $t\in\mathcal{I}(x_0)$. In addition to the assumptions of Theorem \ref{thm:LCBF}, suppose there exists a positive, non-decreasing function $\ell_{gk}\,:\,\mathbb{R}_{\geq0}\rightarrow\mathbb{R}_{\geq0}$ such that $\|g(x)k(x,t)\|\leq\ell_{gk}(\|x\|)\|x\|$ for all $x\in\mathcal{C}$ and $t\in\mathcal{I}(x_0)$, and $\lim_{x\rightarrow\partial\mathcal{C}}\ell_{gk}(\|x\|)\|x\|<\infty$.} Then, the controller
\begin{equation}\label{eq:u_safe}
    u=k(x,t) + k_b(x),
\end{equation}
where $k_b$ is defined as in \eqref{eq:kb}, renders $\mathcal{\text{Int}}(\mathcal{C})$ forward invariant for \eqref{eq:dyn} and ensures the origin is an equilibrium point for the closed-loop system \eqref{eq:dyn}.
\end{corollary}
    
\begin{pf}
    The proof of the first part follows from redefining the drift as $f(x,t)\coloneqq f(x) + g(x)k(x,t)$ and invoking Theorem \ref{thm:LCBF}. The second part follows from noting that $B$ is positive semi-definite on $\text{Int}(\mathcal{C})$, hence $\nabla B(0)=0$ and $k_b(0)=0$. It follows from the assumption that $f(0)=0$ and $k(0,t)=0,\;\forall t\in\mathcal{I}(x_0)$ that $\dot{x}=f(0)+g(0)u(0,t)=0,\;\forall t\in\mathcal{I}(x_0)$.  Hence, the origin is an equilibrium point for the closed-loop system. \qed
\end{pf}

\begin{remark}\label{remark:CBF}
If Assumption \ref{assumption:dyn} fails to hold in the sense that there exists a set $\mathcal{G}\subset\mathcal{N}(\partial\mathcal{C})$ for which $\|\nabla b(x)g(x)\|=0$ for all $x\in\mathcal{G}$, safety can still be guaranteed if $b$ is a \emph{valid} CBF. For Theorem \ref{thm:LCBF}, this follows from the observation that \eqref{eq:CBF2} implies $\|\nabla b(x)g(x)\|=0\implies \nabla b(x)f(x)\leq \alpha_3(h(x))$, which implies the unforced dynamics exhibit a certain degree of safety when $\|\nabla b(x)g(x)\|=0$.
\end{remark}
Theorem \ref{thm:LCBF} and Corollary \ref{cor:safe_control} illustrate how one can modify an existing policy in a minimally invasive fashion to guarantee safety\footnote{Although these results have been established for one safe set, multiple sets can be considered by defining $B(x)\coloneqq \sum_i B_i(x)$, where each $B_i$ is a LCBF over a set $\mathcal{C}_i$ as in \eqref{eq:safe_set}.}.
Based on the bounds on $\dot{B}$ from the proof of Theorem \ref{thm:LCBF}, choosing lower values of $c_b$ implies that $\|\nabla B(x(t))\|$ must reach higher values (i.e., $h(x(t))$ must achieve lower values) before the trajectory $x(t)$ is ``pushed" away from $\partial\mathcal{C}$. We emphasize that knowledge of the bounds on the dynamics from Assumption \ref{assumption:dyn} is not required for implementation of the safeguarding controller and is made to ensure the existence of control values that render the safe set forward invariant. Knowledge of such bounds, however, can be helpful in the selection of $c_b$.  In theory, one could select an arbitrarily small positive value of $c_b$ so that the influence of the safeguarding controller only becomes dominant on $\partial\mathcal{C}$. In practice, however, this could lead to control inputs with large magnitude that may exceed physical actuator limits and could also lead rapid changes in the magnitude of the control input. Hence, $c_b$ is a design parameter that must be carefully selected by the user based upon the specific problem under consideration to ensure a desirable response of the closed-loop system.

\section{Approximate Dynamic Programming}\label{sec:ADP}

\subsection{Infinite-Horizon Nonlinear Optimal Control}
In this section, we shift our attention to the problem of finding a stabilizing control policy for \eqref{eq:dyn} that can be combined with \eqref{eq:kb} to address Problem \ref{prob:main}. {Popular approaches to solving such a problem involve uniting CBFs and CLFs to achieve dual objectives of stability and safety, yet finding a CLF for a general nonlinear control affine system \eqref{eq:dyn} is a non-trivial problem - especially if the dynamics \eqref{eq:dyn} are unknown}. A general way to search for a CLF is to find a control policy that minimizes the infinite-horizon cost functional
\begin{equation}\label{eq:J}
    J(x,u(\cdot))\coloneqq\int_{t_0}^{\infty}\underbrace{\left({x(\tau)^TQx(\tau) + u(\tau)^TRu(\tau)} \right)}_{r(x(\tau),u(\tau))}d\tau,
\end{equation}
where $Q\in\mathbb{R}^{n\times n}$ is positive definite and $R\in\mathbb{R}^{m\times m}$ is symmetric and positive definite.  Solutions to such an optimal control problem are typically characterized in terms of the \emph{optimal value function}
\begin{equation}\label{eq:V*}
    V^*(x(t))\coloneqq{\inf_{u(\cdot)\in \mathcal{U}}}\int_{t}^{\infty}r(x(\tau),u(\tau))d\tau,
\end{equation}
where {$\mathcal{U}$ is the set of admissible control signals}\footnote{{Given an initial condition $x(t_0)\in\mathbb{R}^n$, a control signal $u\,:\,\mathbb{R}_{\geq t_0}\rightarrow\mathbb{R}^m$ is said to be admissible if it is bounded, { piecewise continuous}, and $J(x(t_0),u)$ is finite.}}. Provided $V^*$ is continuously differentiable, it can be shown to be the unique positive definite solution to the \emph{Hamilton-Jacobi-Bellman} (HJB) equation
\begin{equation}\label{eq:HJB}
    0=\inf_{u\in\mathbb{R}^m}\{\nabla V^*(x)f(x) + \nabla V^*(x)g(x)u + r(x,u) \},
\end{equation}
for all $x\in\mathbb{R}^n$ with a boundary condition of $V^*(0)=0$. Provided there exists a continuously differentiable positive definite function $V^*$ satisfying the HJB, taking the minimum on the right-hand side of \eqref{eq:HJB} yields the optimal feedback control policy as
\begin{equation}\label{eq:k^*}
    {k^*(x)=-\frac{1}{2}R^{-1}g(x)^T\nabla V^*(x)^T.}
\end{equation}
\begin{assumption}\label{assumption:value}
There exists a continuously differentiable positive definite function $V^*(x)$ satisfying \eqref{eq:HJB}. Moreover, $\nabla V^*(x)$ is locally Lipschitz.
\end{assumption}
Under Assumption \ref{assumption:value}, the closed-loop system \eqref{eq:dyn} with $u=k^*(x)$ can be shown to be asymptotically stable with respect to the origin using $V^*$ as a CLF \cite[Ch. 8.5]{Sontag}, and Corollary \ref{cor:safe_control} can be used to endow \eqref{eq:k^*} with safety guarantees. Although this approach provides a general way of constructing a CLF for \eqref{eq:dyn}, the practicality of it is hindered by the need to solve the HJB equation \eqref{eq:HJB} for $V^*$, which generally does not admit a closed-form solution, and would require an accurate system model, which may be unavailable in practice. The remainder of this paper is hence dedicated to developing a MBRL framework that can be combined with the results of Sec. \ref{sec:LCBF} to safely learn $V^*$ and the dynamics \eqref{eq:dyn} online.

\subsection{Value Function Approximation}\label{sec:VFA}
Since the optimal value function $V^*$ is unknown and difficult to compute in general, we seek a parametric approximation of $V^*$ over some compact set $\chi\subset\mathbb{R}^n$ containing the origin. To this end, we leverage state following (StaF) kernels introduced in \cite{KamalapurkarAutomatica16-staf,DixonTNNLS19} to generate a local approximation of the value function within a smaller compact set $\mathcal{B}_{l}(x)\subset\chi$ that follows the system trajectory. Using StaF kernels, the value function \eqref{eq:V*} can be represented at points $y\in\mathcal{B}_{l}(x)$ as \cite{KamalapurkarAutomatica16-staf}
\begin{equation}\label{eq:V^*_staf}
    V^*(y)=W(x)^T\phi(y,c(x)) + \varepsilon(x,y),
\end{equation}
where $W\,:\,\chi\rightarrow\mathbb{R}^L$ is a continuously differentiable ideal weight function, $\phi\,:\,\chi\times\chi\rightarrow\mathbb{R}^L$ is a vector of $L\in\mathbb{N}$ bounded, positive definite, and smooth kernel functions, where $c(x)=[c_1(x)\,\dots\,c_L(x)]^T$ and $c_i(x)\in\mathcal{B}_l(x),\,i\in\{1,\dots,L\}$ denotes the center of the $i$th kernel, and $\varepsilon\,:\,\chi\times\chi\rightarrow\mathbb{R}$ denotes the function reconstruction error. {Further details regarding the selection} {of kernel functions can be found in \cite[Footnote 7]{KamalapurkarAutomatica16-staf} and \cite{DixonTNNLS19}}. The ideal weight function $W$ in \eqref{eq:V^*_staf} is generally unknown, and is therefore replaced with approximations\footnote{The use of separate weight estimates is motivated by the fact that \eqref{eq:BE_original} is affine in $\hat{W}_c$ \cite{VamvoudakisAutomatica10}.} $\hat{W}_c(t),\,\hat{W}_a(t)\in\mathbb{R}^L$ yielding the approximate value function and approximate optimal control policy as
\begin{subequations}\label{eq:staf_approx}
\begin{equation}\label{eq:V_staf}
    \hat{V}(y,x,\hat{W}_c)\coloneqq{\hat{W}_c^T\phi(y,c(x))},
\end{equation}
\begin{equation}\label{eq:k_staf}
    \hat{k}(y,x,\hat{W}_a)\coloneqq-\frac{1}{2}R^{-1}g(y)^T\nabla\phi(y,c(x))^T\hat{W}_a.
\end{equation}
\end{subequations}

\subsection{System Identification}
In addition to not knowing the value function, we now assume that the system drift from \eqref{eq:dyn} is unknown, but can be expressed as a linear combination of weights $\theta\in\mathbb{R}^p$ and user-defined basis functions $Y\,:\,\mathbb{R}^n\rightarrow\mathbb{R}^{n\times p}$ such that over a given compact set $\chi\subset\mathbb{R}^n$ 
\begin{equation}\label{eq:f_approx}
    f(x)=Y(x)\theta + \varepsilon_{\theta}(x),
\end{equation}
where $\varepsilon_{\theta}\,:\,\mathbb{R}^n\rightarrow\mathbb{R}^n$ is the bounded \emph{unknown} function reconstruction error. The basis $Y$ may capture physical knowledge of the dynamics or may represent user-defined basis functions, such as polynomials, radial basis functions, and pre-trained neural networks with tunable outer layer weights, that can approximate functions arbitrarily closely over compact sets. As the ideal weights\footnote{For a given basis $Y$, the ideal weight vector is defined as $\theta\coloneqq\mathrm{arg\,min}_{\hat{\theta}\in\mathbb{R}^p}\sup_{x\in\chi}\|f(x) - Y(x)\hat{\theta}\|$ (see \cite[Ch. 8.7]{IoannouFidan}).} for the given basis are assumed to be unknown, let $\hat{\theta}\in\mathbb{R}^p$ denote an estimate of $\theta$ and let $\hat{f}(x,\hat{\theta})\coloneqq Y(x)\hat{\theta}$ denote the corresponding approximated drift. Provided $\hat{\theta}$ is updated by a specific class of parameter identifiers\footnote{See \cite[Assumption 2]{KamalapurkarAutomatica16-mbrl} for conditions that such an identifier must satisfy and \cite{Chowdhary,DixonIJACSP19,DeptulaAutomatica20,DeptulaAutomatica21} for examples of identifiers. \label{footnote:sysID}}, then {one can establish the existence of a Lyapunov-like function $V_{\theta}\,:\,\mathbb{R}^p\times\mathbb{R}_{\geq0}\rightarrow\mathbb{R}_{\geq0}$ satisfying 
\begin{equation}\label{eq:Vtheta}
    \dot{V}_{\theta}(\tilde{\theta},t)\leq -K_{\theta}\|\tilde{\theta}\|^2 + D_{\theta}\|\tilde{\theta}\|,
\end{equation}
for all $\tilde{\theta}\in\mathbb{R}^p$ and $t\in\mathbb{R}_{\geq0}$, where $\tilde{\theta}\coloneqq\theta - \hat{\theta}$ denotes the weight estimation error and $K_{\theta},D_{\theta}\in\mathbb{R}_{>0}$ are positive constants, the latter of which depends upon $\varepsilon_{\theta}$ (cf. \cite{KamalapurkarAutomatica16-mbrl,KamalapurkarAutomatica16-staf}). The bound in \eqref{eq:Vtheta} implies that, under such an identification scheme, the weight estimation error exponentially decays to a ball about the origin, the size of which also depends upon $\varepsilon_{\theta}$.}

\subsection{Bellman Error}
A performance metric for learning the ideal parameters of the value function and policy can be obtained by replacing the optimal value function, optimal policy, and true drift in the HJB \eqref{eq:HJB} with their corresponding approximations yielding the \emph{Bellman error} (BE)
\begin{equation}
\begin{aligned}\label{eq:BE_original}
\delta(y,x,\hat{W}_c,\hat{W}_a,\hat{\theta})\coloneqq & \nabla \hat{V}(y,x,\hat{W}_c)\hat{f}(y,\hat{\theta}) \\ & +\nabla \hat{V}(y,x,\hat{W}_c)g(y)\hat{k}(y,x,\hat{W}_a) \\ & 
+ r(y,\hat{k}(y,x,\hat{W}_a)).
\end{aligned}
\end{equation}
Given weight estimates $(\hat{W}_c,\,\hat{W}_a)$, the BE, evaluated at any point in the state-space, encodes the ``distance" of the approximations $(\hat{V},\,\hat{k})$ from their true values $(V^*,\,k^*)$. Thus, the objective is to select $(\hat{W}_c,\,\hat{W}_a)$ such that $\delta(y,x,\hat{W}_c,\hat{W}_a,\hat{\theta})=0$ for all $x\in\chi$ and all $y\in \mathcal{B}_l(x)$, which is accomplished by updating $(\hat{W}_c,\,\hat{W}_a)$ online using techniques from adaptive control \cite{Krstic,Chowdhary,VamvoudakisAutomatica10}.

\section{Safe Exploration via Simulation of Experience}\label{sec:learning}
{This section builds upon the approach from \cite{KamalapurkarAutomatica16-mbrl,KamalapurkarAutomatica16-staf,Kamalapurkar} to develop a safe exploration framework for learning $V^*$ and $k^*$ online while guaranteeing safety. To this end, define the control policy for \eqref{eq:dyn} as
\begin{equation}\label{eq:u_ADP}
    u=\hat{k}(x,x,\hat{W}_a) -\frac{c_b}{2}R^{-1}g(x)^T\nabla B(x)^T,
\end{equation}
where $\hat{k}$ is defined as in \eqref{eq:k_staf}, $c_b\in\mathbb{R}_{>0}$ is a gain, and $\nabla B$ is a LCBF as in \eqref{eq:B}, which is a special case of the controller proposed in Corollary \ref{cor:safe_control}.} Define the regressor $\omega(t)\coloneqq \nabla\phi(x(t),c(x(t)))\cdot(\hat{f}(x(t),\hat{\theta}(t)) + g(x(t))u(t))$ and the resulting BE along the system trajectory as
\begin{equation}\label{eq:BE}
    \delta_t(t)\coloneqq r(x(t),u(t)) + \hat{W}_c(t)^T\omega(t).
\end{equation}
In contrast to \eqref{eq:BE_original}, the version of the BE in \eqref{eq:BE} is a function of $\nabla B$ and therefore selecting the weight estimates to minimize \eqref{eq:BE} may not correspond to the minimization of the original BE in \eqref{eq:BE_original}. That is, even if $(\hat{W}_c,\,\hat{W}_a)\rightarrow W$, the BE in \eqref{eq:BE} may be large at certain points in the state-space because of the influence of the safeguarding component of \eqref{eq:u_ADP}, making \eqref{eq:BE} a non-ideal performance metric for learning. {However, given an approximate model of the system}, the {BE can be evaluated at any point in the state-space \cite{KamalapurkarAutomatica16-mbrl} using a \emph{different} policy to generate data more representative of \eqref{eq:BE_original}}.  To facilitate this approach, define the family of mappings $\{x_i\,:\,\chi\times\mathbb{R}_{\geq t_{0}}\rightarrow\chi \}_{i=1}^N$ such that each $x_i(x(t),t)\in\mathcal{B}_l(x(t))$ maps the current state $x(t)$ to some unexplored point in $\mathcal{B}_{l}(x(t))$. For each extrapolated trajectory\footnote{ See \cite[Rem. 5]{DeptulaAutomatica21} for details on generating trajectories.\label{footnote:extrap}}, $t\mapsto x_i(x(t),t),\,i\in\{1,\dots,N\}$, {we define an \emph{exploratory policy} as $u_i(t)\coloneqq \hat{k}(x_i(x(t),t),x(t),\hat{W}_a(t))$ with $\hat{k}$ as in \eqref{eq:k_staf}, which yields the BE along the extrapolated system trajectories as}\footnote{Mappings with the subscript $i$ denote evaluations along the extrapolated trajectories.}
\begin{equation}\label{eq:BE_extrap}
    \delta_i(t)\coloneqq r(x_i(x(t),t),u_i(t)) + \hat{W}_c(t)^T\omega_i(t).
\end{equation}
Note that the exploratory policy $u_i$ is not augmented with the safeguarding controller, thereby allowing for maximum exploration of the state-space by the extrapolated system trajectories without risking safety violation of the original system. Consequently, the BE in \eqref{eq:BE_extrap} is representative of the original BE from \eqref{eq:BE_original} and is used to update the weight estimates using a recursive least squares update law as
\begin{equation}\label{eq:Wc_dot}
    \dot{\hat{W}}_c(t)= -\Gamma(t)\left(k_{c1}\frac{\omega(t)}{\rho^2(t)}\delta_t(t) + \frac{k_{c2}}{N}\sum_{i=1}^N\frac{\omega_i(t)}{\rho^2_i(t)}\delta_i(t) \right)
\end{equation}
\begin{equation}\label{eq:Gamma_dot}
    \dot{\Gamma}(t)=\beta_c\Gamma(t) - \Gamma(t)\left(k_{c1}\Lambda(t) + \frac{k_{c2}}{N}\sum_{i=1}^N\Lambda_i(t) \right)\Gamma(t),
\end{equation}
where $\Lambda(t)\coloneqq\tfrac{\omega(t)\omega(t)^T}{\rho^2(t)}$, $\rho(t)\coloneqq 1 + \gamma_c\omega(t)^T\omega(t)$ is a normalizing signal with a gain of $\gamma_c\in\mathbb{R}_{>0}$, $k_{c1},\,k_{c2}\in\mathbb{R}_{>0}$ are learning gains, and $\beta_c\in\mathbb{R}_{>0}$ is a forgetting factor. {In \eqref{eq:Wc_dot}-\eqref{eq:Gamma_dot} the terms $\omega_i,\rho_i,\Lambda_i$ are defined in a similar manner to $\omega,\rho,\Lambda$.} The weight update law for $\hat{W}_a$ is then selected as
\begin{equation}\label{eq:Wa_dot}
\begin{aligned}
    \dot{\hat{W}}_a(t)=&{\text{proj}\bigg\{}-k_{a1}(\hat{W}_a(t)-\hat{W}_c(t)) - k_{a2}\hat{W}_a(t)
    \\&+ \frac{k_{c1}}{4\rho^2(t)}{G_{\phi}(t)}^T\hat{W}_a(t)\omega(t)^T\hat{W}_c(t) \\&+ \sum_{i=1}^{N}\frac{k_{c2}}{4N\rho^2_i(t)}{G_{\phi,i}(t)}^T\hat{W}_a(t)\omega_i(t)^T\hat{W}_c(t){\bigg\}},
\end{aligned}
\end{equation}
where {$G_{\phi}\coloneqq \nabla \phi(x,c(x))G_R(x)\nabla\phi(x,c(x))^T$,  $G_R\coloneqq g(x)R^{-1}g(x)^T$},  $k_{a1},\,k_{a2}\in\mathbb{R}_{>0}$ are learning gains, and  {proj$\{\cdot\}$ is a smooth projection operator, standard in the adaptive control literature \cite[Appendix E]{Krstic}, that ensures the weight estimates remain bounded}. {The following proposition shows that the policy in \eqref{eq:u_ADP} renders $\text{Int}(\mathcal{C})$ forward invariant for the closed-loop system \eqref{eq:dyn}.}

\begin{proposition}\label{prop:ADP_safe}
Consider system \eqref{eq:dyn}, a set $\mathcal{C}\subset\mathbb{R}^n$ as in \eqref{eq:safe_set} with $0\in\text{Int}(\mathcal{C})$, { and let $b\,:\,\text{Int}(\mathcal{C})\rightarrow\mathbb{R}_{\geq0}$ be a candidate CBF. Provided Assumptions \ref{assumption:dyn}-\ref{assumption:value} hold and there exists a positive, non-decreasing function $\ell_{g\phi}\,:\,\mathbb{R}_{\geq0}\rightarrow\mathbb{R}_{\geq0}$ such that $\|g(x)^T\nabla\phi(x,c(x))^T\|\leq\ell_{g\phi}(\|x\|)\|x\|$ for all $x\in\mathcal{C}$ and $\lim_{x\rightarrow\partial\mathcal{C}}\ell_{g\phi}(\|x\|)\|x\|<\infty$}, then the control policy in \eqref{eq:u_ADP} and weight update law in \eqref{eq:Wa_dot} ensure $\text{Int}(\mathcal{C})$ is forward invariant for \eqref{eq:dyn} and ensure the origin is an equilibrium point for the closed-loop system \eqref{eq:dyn}.
\end{proposition}

\begin{pf}
    {Under the assumptions of the proposition, the nominal policy $\hat{k}(x,x,\hat{W}_a)$ can be bounded as $\|\hat{k}(x,x,\hat{W}_a)\|\leq\tfrac{1}{2}\lambda_{\max}(R^{-1})\bar{W}_a\ell_{g\phi}(\|x\|)\|x\|$ for all $x\in \mathcal{C}$, where $\|\hat{W}_a\|\leq\bar{W}_a$ for some $\bar{W}_a\in\mathbb{R}_{>0}$ follows from the use of the projection operator in \eqref{eq:Wa_dot}.} The proof then follows from letting $k(x,t)=\hat{k}(x,x,\hat{W}_a(t))$ and invoking Corollary \ref{cor:safe_control}. \qed
\end{pf}
    
In contrast to related approaches \cite{VamvoudakisIJRNC20,GreeneLCSS20,KamalapurkarACC21,CohenCDC20,KiumarsiIJRNC21}, the above proposition does not make use of any Lyapunov-based arguments that require the value function to decrease along the system trajectory, only that $\hat{W}_a$ remains bounded, which is guaranteed by the use of  the projection operator in \eqref{eq:Wa_dot}. The following assumption outlines the exploration conditions required for the approximations to converge to a neighborhood of their ideal values.

\begin{assumption}[\cite{KamalapurkarAutomatica16-staf,DeptulaAutomatica20}]\label{assumption:PE}
    There exist constants $T\in\mathbb{R}_{>0}$ and $\underline{c}_1,\,\underline{c}_2,\,\underline{c}_3\in\mathbb{R}_{\geq0}$ such that for all $t\geq t_0$ we have $\underline{c}_1I_L\leq\int_{t}^{t+T} \Lambda(\tau)d\tau$, $\underline{c}_2I_L\leq\int_{t}^{t+T}\tfrac{1}{N}\sum_{i=1}^N\Lambda_i(\tau)d\tau$, and $\underline{c}_3I_L\leq\inf_{t\geq t_0}(\frac{1}{N}\sum_{i=1}^N\Lambda_i(t))$, where at least one of $\underline{c}_1,\,\underline{c}_2,\underline{c}_3$ is strictly greater than zero\footnote{The first condition in Assumption \ref{assumption:PE} is the persistence of excitation (PE) condition, whose satisfaction cannot be verified in general, but can be achieved heuristically by including an exploration/probing signal in the control input. The second condition is the same PE condition, but is placed on the \emph{extrapolated trajectories}. Since the extrapolated trajectories are user-defined through the mappings $\{x_i(x(t),t) \}_{i=1}^N$, {one can construct these mappings in an attempt to satisfy such a condition \cite[Rem. 5]{DeptulaAutomatica21} without having to excite the original system.} The third condition can be heuristically satisfied by selecting many extrapolation points $N$; however, this scales poorly with the state dimension (cf. \cite{KamalapurkarAutomatica16-staf}).\label{remark:PE}}.
\end{assumption}

\subsection{Stability Analysis}
To facilitate the analysis of the closed-loop system under the influence of the controller in \eqref{eq:u_ADP}, {define the weight estimation errors $\tilde{W}_c(t)\coloneqq W(x(t))-\hat{W}_c(t)$, $\tilde{W}_a(t)\coloneqq W(x(t))-\hat{W}_a(t)$, and a composite state vector $Z\coloneqq[x^T,\,\tilde{W}_c^T,\,\tilde{W}_a^T,\,\tilde{\theta}^T]^T$. Now consider the following Lyapunov function candidate
\begin{equation*}
    V_L(Z,t)\coloneqq V^*(x) + \frac{1}{2}\tilde{W}_c^T\Gamma^{-1}(t)\tilde{W}_c + \frac{1}{2}\tilde{W}_a^T\tilde{W}_a + V_{\theta}(\tilde{\theta},t),
\end{equation*}
where $V_{\theta}$ is from \eqref{eq:Vtheta}.} If $\lambda_{\min}(\Gamma^{-1}(t_0))>0$ and Assumption \ref{assumption:PE} is satisfied, then $\Gamma(t)$ can be shown to satisfy $\underline{\Gamma}I_L\leq \Gamma(t)\leq \overline{\Gamma}I_L$ for all $t\geq t_0$, where $\underline{\Gamma},\,\overline{\Gamma}\in\mathbb{R}_{>0}$ \cite[Lemma 1]{KamalapurkarAutomatica16-staf}, which implies $V_L$ is positive definite and hence satisfies $\eta_1(\|Z\|)\leq V_L(Z,t)\leq \eta_2(\|Z\|)$ for all $t\geq t_0$ with $\eta_1,\,\eta_2\in\mathcal{K}$ \cite[Lemma 4.3]{Khalil}. The following theorem illustrates that the safe exploration scheme ensures the system state and weight estimation errors remain uniformly ultimately bounded. 

\begin{theorem}\label{thm:UUB}
    Consider system \eqref{eq:dyn}, the cost functional in \eqref{eq:J}, a set $\mathcal{C}\subset\mathbb{R}^n$ as in \eqref{eq:safe_set} with $0\in\text{Int}(\mathcal{C})$, and {let $b\,:\,\text{Int}(\mathcal{C})\rightarrow\mathbb{R}_{\geq0}$ be a candidate CBF for \eqref{eq:dyn} on $\mathcal{C}$}. Let the optimal value function and corresponding optimal control policy be approximated over a compact set $\chi\subset\mathbb{R}^n$ as detailed in Sec. \ref{sec:VFA}, and let $\mathcal{B}_\zeta(0)\in\chi\times\mathbb{R}^{2L+p}$ be a closed ball of radius $\zeta\in\mathbb{R}_{>0}$ centered at $Z=0$. Provided Assumptions \ref{assumption:dyn}-\ref{assumption:PE}, {the conditions of Proposition \ref{prop:ADP_safe}, and the inequaility in \eqref{eq:Vtheta}} hold, $x(t_0)\in\text{Int}(\mathcal{C})$, and
    \begin{equation}\label{eq:suff_cond}
    \lambda_{\min}(M)>0,\quad\sqrt{\tfrac{2\iota}{\kappa}} < \eta_1^{-1}\left(\eta_2(\zeta)\right),
    \end{equation}
    where {
    \begin{equation}
        M\coloneqq
        \begin{bmatrix}
            \frac{\underline{c}k_{c2}}{4} & -\frac{\varphi_{ac}}{2} & -\frac{\varphi_{c\theta}}{2} \\
            -\frac{\varphi_{ac}}{2} & \left(\frac{k_{a1} + k_{a2}}{4} - \varphi_a \right) & 0 \\
            -\frac{\varphi_{c\theta}}{2} & 0 & \tfrac{K_{\theta}}{4}
        \end{bmatrix},
    \end{equation}}
    with $\underline{c}\coloneqq(\frac{\underline{c}_{3}}{2} + \frac{\beta_c}{2k_{c2}\overline{\Gamma}})$ and $\kappa,\,\iota,\,\varphi_a,\,\varphi_{ac},{\varphi_{c\theta}}\in\mathbb{R}_{>0}$ defined in the Appendix, then the control policy in \eqref{eq:u_ADP} and weight update laws in \eqref{eq:Wc_dot}, \eqref{eq:Gamma_dot}, \eqref{eq:Wa_dot} guarantee that $\text{Int}(\mathcal{C})$ is forward invariant for the closed-loop system \eqref{eq:dyn} and that $Z(t)$ is uniformly ultimately bounded such that
    \begin{equation}\label{eq:UUB}
        \limsup_{t\rightarrow\infty} \|Z(t)\|\leq \eta_1^{-1}\circ\eta_2(\sqrt{{2\iota}/{\kappa}} ).
    \end{equation}
\end{theorem}
    
\begin{remark}\label{remark:suff_cond}
    It is difficult to verify when the sufficient conditions in \eqref{eq:suff_cond} are satisfied because $M$ and $\iota$ depend on terms that are either unknown, such as $\varepsilon$, or completely determined by the system trajectory, such as $\underline{c}$, and therefore one cannot formally guarantee the satisfaction of \eqref{eq:suff_cond} a priori. Although the stability guarantees depend upon these conditions, the safety guarantees of the proposed framework have already been established in Proposition \ref{prop:ADP_safe}, which is in contrast to related approaches that use the value function as a safety certificate. The definitions provided in the Appendix imply that \eqref{eq:suff_cond} can be satisfied by selecting a sufficient number of basis functions $L$, which decreases the approximation error $\varepsilon$, by choosing $R$ such that $\lambda_{\min} (R)$ is large, and by ensuring that $\underline{c}$ is large, which can be achieved using the methods mentioned in Footnote \ref{remark:PE}. The ultimate bound is also a function of $\|\nabla B(x(t))\|_{\infty}$ and therefore selecting $c_b$ to be small can aid in satisfying the sufficient conditions.
\end{remark}

\begin{pf}
    {The forward invariance of $\text{Int}(\mathcal{C})$ follows from Proposition \ref{prop:ADP_safe}.} Expressing \eqref{eq:BE} and \eqref{eq:BE_extrap} in terms of the weight estimation errors and then subtracting the right-hand side of the HJB equation \eqref{eq:HJB} using the StaF representation of $V^*$ from \eqref{eq:V^*_staf} yields an alternative form of the BEs as $\delta_t = -\tilde{W}_c^T\omega + \frac{1}{4}\tilde{W}_a^TG_\phi \tilde{W}_a { - W^T\nabla\phi Y\tilde{\theta}} - \frac{1}{2}c_b \nabla B G_R\nabla\phi^T\tilde{W}_a + \frac{1}{4}c_b^2\nabla BG_R\nabla B^T + \Delta$ and $\delta_i=-\tilde{W}_c^T\omega_i + \frac{1}{4}\tilde{W}_a^TG_{\phi,i} \tilde{W}_a { - W_i^T\nabla\phi_i Y_i\tilde{\theta}} + \Delta_i$, where functional dependencies have been suppressed for ease of presentation and $\Delta(x),\,\Delta_i(x_i)$ consist of terms that are uniformly bounded over $\chi$. Taking the derivative of $V_L$ along $Z(t)$ yields
    \begin{equation*}
    \begin{aligned}
        \dot{V}_L= &\nabla V^*f + \nabla V^*gu + \frac{1}{2}\tilde{W}_c^T\Gamma^{-1}(\dot{W}-\dot{\hat{W}}_c) \\ 
        &- \frac{1}{2}\tilde{W}_c^T\Gamma^{-1}\dot{\Gamma}\Gamma^{-1}\tilde{W}_c + \frac{1}{2}\tilde{W}_a^T(\dot{W}-\dot{\hat{W}}_a) { + \dot{V}_{\theta}}.
    \end{aligned}
    \end{equation*}
    {Using $\nabla V^*f = -\nabla V^*gk^* - k^{*T}Rk^* - x^TQx$ from the HJB \eqref{eq:HJB}}, $\dot{W}=\nabla W(f+gu)$, substituting in the weight update laws \eqref{eq:Wc_dot}, \eqref{eq:Gamma_dot}, \eqref{eq:Wa_dot} using the alternate form of the BEs, expressing everything in terms of the weight estimation errors, upper bounding and then completing squares yields $\dot{V}_L\leq -\kappa\|Z\|^2  - \underline{Z}^TM\underline{Z}+ \iota$, where $\underline{Z}\coloneqq[\|\tilde{W}_c\|\,\|\tilde{W}_a\|{\,\|\tilde{\theta}\|}]^T$. Provided \eqref{eq:suff_cond} holds, further bounding yields $\dot{V}_L\leq -\frac{\kappa}{2}\|Z\|^2$ $\forall\zeta\geq\|Z\|\geq \sqrt{\tfrac{2\iota}{\kappa}}$. Finally, invoking \cite[Thm. 4.18]{Khalil} implies $Z(t)$ is uniformly ultimately bounded such that \eqref{eq:UUB} holds. \qed
\end{pf}

\section{Numerical Examples}\label{sec:sim}

\paragraph*{Nonlinear System}\label{sec:nonlinear_sim}
To demonstrate the efficacy of the developed approach, we apply our method to the scenario from \cite{JankovicAutomatica18}. Consider a system as in \eqref{eq:dyn} with $x\in\mathbb{R}^2,\,u\in\mathbb{R}$, $f(x)=[-0.6x_1-x_2,\,x_1^3]^T$ and $g(x)=[0,x_2]^T$. The drift is assumed to be unknown but linear in the unknown parameters (i.e., $\varepsilon_{\theta}\equiv 0$) and is represented as $f(x)=Y(x)\theta$, where $Y(x)=[x_1,\,x_2,\,0;\,0,\,0,\,x_1^3]$ and $\theta=[-0.6,\,-1,1]^T$. The objective is to drive $x(t)$ to the origin while ensuring $x(t)$ remains in a set $\mathcal{C}$ as in \eqref{eq:safe_set} defined by $h(x)=px_2^2-x_1+1$, where $p\in\{-1,1\}$ determines if $\mathcal{C}$ is convex ($p=-1$) or not ($p=1$). To the best of our knowledge, such a set cannot be rendered safe using the approaches from \cite{VamvoudakisIJRNC20,GreeneLCSS20,KamalapurkarACC21} as the constraints are not of the form $\underline{a}_i<x_i<\overline{a}_i\;\forall i\in\{1,\dots,n\}$ with $\underline{a}_i,\,\overline{a}_i\in\mathbb{R}$. For each simulation, a safeguarding controller is obtained by constructing the CBF $b(x)=1/h(x)$ and then constructing an LCBF as in \eqref{eq:B}. To obtain a stabilizing control policy, we define an optimal control problem as in \eqref{eq:J} with $Q=I_2$ and {$R=1$}. The resulting value function is approximated using a basis of $L=3$ StaF kernels $\phi(x,c(x))=[\phi_1(x,c_1(x))\,\phi_2(x,c_2(x))\,\phi_3(x,c_3(x))]^T$, where each kernel is selected as the polynomial kernel $\phi_i(x,c_i(x))=x^Tc_i(x)$. The centers of each kernel are placed on the vertices of an equilateral triangle centered at the current state as $c_i(x)=x+\nu(x)d_i$, where $d_i\in\mathbb{R}^2$ corresponds to vertices of the triangle and $\nu(x)\coloneqq\tfrac{x^Tx}{x^Tx + 1}$. The learning gains are selected as $k_{c1}=0.1,\,k_{c2}=1,\,k_{a1}=1,\,k_{a2}=0.1,\,\gamma_c=1,\,\beta_c=0.001$ and the weights are initialized as $\hat{W}_c(t_0)=\hat{W}_a(t_0)=[0.5,\,0.5,\,0.5]^T$ and $\Gamma(t_0)=100I_3$. The learning procedure outlined in Sec. \ref{sec:learning} is carried out by extrapolating the BE to one point from a uniform distribution over a $\nu(x)\times\nu(x)$ square centered at $x$ every time-step, where the drift parameters are identified online using integral concurrent learning \cite{DixonIJACSP19,DeptulaAutomatica21}.

\begin{figure}[t]
    \centering
    \includegraphics{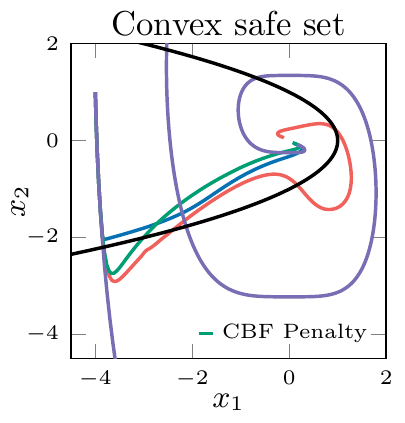}\hfill %
    \includegraphics{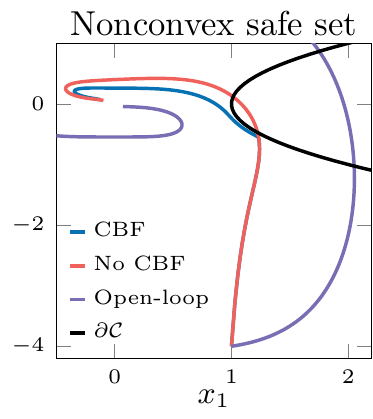}
    \caption{Trajectories of the nonlinear system evolving in the convex (left) and nonconvex (right) safe sets. In each plot the blue curve denotes the trajectory under the RL policy augmented with the safeguarding controller, the orange curve denotes the trajectory under only the RL policy, the green curve denotes the trajectory under the RL policy with a cost function of $r(x,u)=\|x\|^2 + u^2 + 20B(x)$, the purple curve denotes the open-loop trajectory, and the black curve denotes the boundary of the safe set.}
    \label{fig:nonlinear_traj}
\end{figure}

\begin{figure}[ht]
    \centering
    \includegraphics{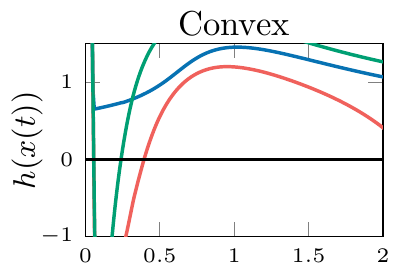}\hfill %
    \includegraphics{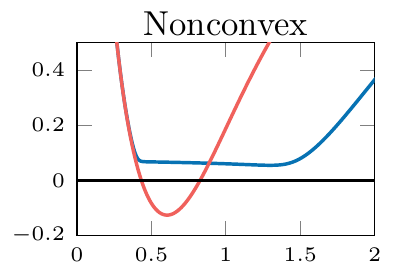}

    \includegraphics{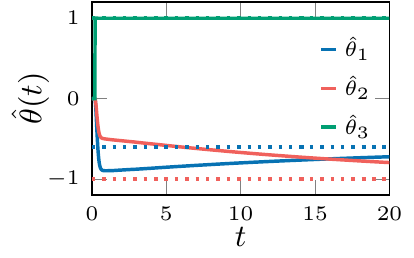} \hfill %
    \includegraphics{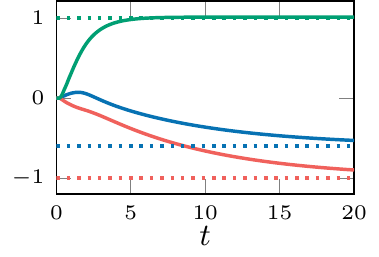}
    \caption{Evolution of the CBF $h$ along the system trajectory (top) and trajectory of the estimated drift dynamics weights along the trajectory of the nonlinear system under the RL policy augmented with the safeguarding controller (bottom). In the top plots, the curves of various color have the same interpretation as those in Fig. \ref{fig:nonlinear_traj}. In the bottom plots, the dotted lines of corresponding color denote the true values of the uncertain parameters.}
    \label{fig:nonlinear_extra}
\end{figure}

To illustrate efficacy of the developed approach, the system is simulated with the safeguarding controller, without the safeguarding controller, { and without the safeguarding controller where the LCBF is incorporated into the cost function in a similar fasion to \cite{CohenCDC20,KiumarsiIJRNC21}.} For the first simulation, we consider the convex safe set defined by $p=-1$ and the policy in \eqref{eq:u_ADP} with $c_b=1$, the results of which are provided in Fig. \ref{fig:nonlinear_traj} (left) and Fig. \ref{fig:nonlinear_extra} (left). As shown in Fig. \ref{fig:nonlinear_traj}, the RL policy in \eqref{eq:u_ADP} augmented with the safeguarding controller stabilizes $x(t)$ to the origin without leaving the safe set. { Without the safeguarding component, the RL policy stabilizes the system, but violates the safety constraints multiple times. When the CBF term is included in the problem's cost function, the system initially violates the safety constraints, but eventually converges to a safe policy. For this particular example, the system rapidly approaches the boundary of the safe set (see the top left plot in Fig. \ref{fig:nonlinear_extra}) and crosses into the unsafe region before enough time has passed for convergence to a suitable policy without additional help from the safeguarding controller. This phenomenon underscores the crucial distinction between the approach taken herein and those in, e.g., \cite{CohenCDC20,KiumarsiIJRNC21}, where safety is enforced by including a CBF in the problem's cost function. In essence, these aforementioned approaches aim to \emph{learn a safe policy}, whereas the approach presented herein aims to \emph{safely learn a performance policy}. Ultimately, this allows for safe exploration while learning an uncertain system model (Fig. \ref{fig:nonlinear_extra}, bottom) and an approximately optimal policy.} Moreover, note that the safeguarding controller is minimally invasive in the sense that it intervenes only when absolutely necessary to prevent safety violation. In fact, the trajectories of the controlled system with and without the safeguarding controller are almost identical up until the point at which $x(t)$ approaches $\partial\mathcal{C}$. To demonstrate the ability of the RL policy to safely stabilize the system within a non-convex safe set, a second simulation under the RL policy with and without the safeguarding controller is run with $p=1$, the results of which are shown in Fig. \ref{fig:nonlinear_traj} (right) and Fig. \ref{fig:nonlinear_extra} (right). All parameters remain the same as before with the exception of $c_b=0.001$ and $\Gamma(t_0)=10I_3$. Once again, the policy in \eqref{eq:u_ADP} renders the system safe and the trajectory under the nominal and safe policy only diverge from each other if intervention from the safeguarding controller is required to ensure safety.

\paragraph*{Collision Avoidance}
We now examine a simple scenario to demonstrate some interesting properties of the proposed method. Consider a mobile robot modeled as a two-dimensional single integrator $\dot{x}=u$ tasked with navigating to the origin while avoiding a circular obstacle centered at $x_o\in\mathbb{R}^2$ with a radius of $r_o\in\mathbb{R}_{>0}$. The safety objective can be addressed by considering a set $\mathcal{C}$ defined by $h(x)=\|x-x_o\|^2 - r_o^2$. Similar to the previous example, the safe set is not a hyper-rectangle and therefore the techniques in \cite{VamvoudakisIJRNC20,GreeneLCSS20,KamalapurkarACC21} cannot be applied. To obtain a stabilizing control policy we associate with the single integrator an optimal control problem as in \eqref{eq:J} with $Q=I_2$ and $R=I_2$. Since the system is linear and the cost is quadratic, one could solve the algebraic Ricatti equation (ARE) to obtain the optimal policy and then augment it with the safeguarding controller as in Corollary \ref{cor:safe_control}. However, as demonstrated in the subsequent numerical results, such an approach presents certain limitations. To compare the analytical solution with the learning-based solution, the corresponding value function is approximated using the same parameters as in the previous example with all weights initialized to 1 and $\Gamma(t_0)=10I_3$. Since the dynamics for this example are trivial, no system identification is performed.

\begin{figure}[ht]
    \centering

    \includegraphics{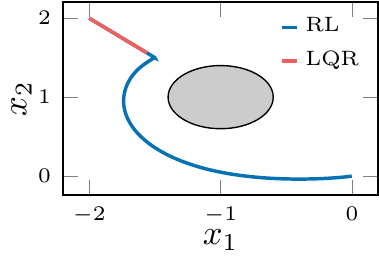}\hfill %
    \includegraphics{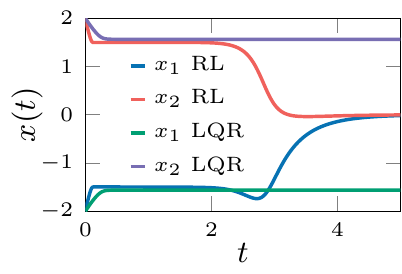}

    \includegraphics{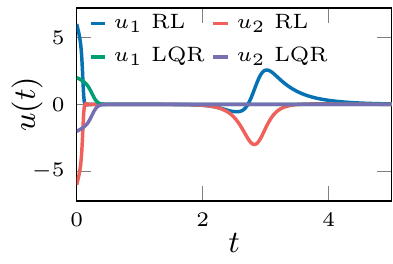} \hfill%
    \includegraphics{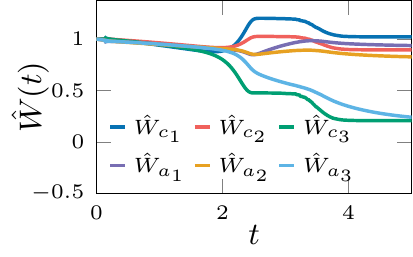}%

    \caption{State (top), control (bottom left), and weight (bottom right) trajectory for the single integrator simulation. In the top left plot the gray disk denotes the obstacle.}
    \label{fig:linear_extra}
\end{figure}

Simulations are performed to compare the policy from \eqref{eq:u_ADP} with that obtained from solving the ARE, both of which are augmented with the safeguarding controller with $c_b=0.1$. The resulting system trajectories are provided in Fig. \ref{fig:linear_extra}, where the trajectory under the RL policy navigates around the obstacle and converges to the origin, whereas the trajectory under the linear quadratic regulation (LQR) policy gets stuck behind the obstacle. Note that LQR policy augmented with the safeguarding controller is a continuous \emph{time-invariant} feedback controller and can therefore not achieve dual objectives of obstacle avoidance and stability from certain initial conditions\footnote{The unfamiliar reader is referred to \cite[Ch. 4.1]{Liberzon-switched-systems} for an intuitive discussion on this topic.}. On the other hand, the RL policy is continuous, but is also \emph{time-varying} because the policy explicitly depends on the evolution of $\hat{W}_a$. {Indeed, adaptive control methods in general can be seen as a form of nonlinear \emph{dynamic} feedback \cite{Krstic}}. As illustrated in Fig. \ref{fig:linear_extra} (top), both trajectories quickly approach the obstacle and initially get stuck, failing to make progress towards the origin. However, unlike the static LQR policy, the weights of the learning-based controller dynamically evolve, as shown in Fig. \ref{fig:linear_extra} (bottom), and eventually converge to a new policy that successfully navigates the system around the obstacle and to the origin.

\section{Conclusions}
In this paper, we developed a safe MBRL framework that allows one to learn online the value function of an optimal control problem and the drift dynamics of an uncertain control affine system while satisfying safety constraints given as CBFs. Our approach was facilitated by the introduction of a new class of CBFs, termed LCBFs, that were used to augment a learning-based control policy to guarantee stability and safety. The benefits of the proposed method were illustrated by introducing numerical examples that, to the best of our knowledge, cannot be handled by related approaches. Directions for future research include integrating zeroing CBFs into the developed framework using approaches such as in \cite{TaylorACC20,LopezLCSS21}, which may address the limitations mentioned after Corollary \ref{cor:safe_control}, and may strengthen the proposed framework in the context of uncertain systems.

\begin{ack}                               
    We are indebted to the anonymous reviewers, whose insightful comments and constructive criticism have significantly improved the quality of this work. This work was partially supported by the NSF under grants IIS-1723995, IIS-2024606, and DGE-1840990. Any opinions, findings, and conclusions or recommendations expressed in this material are those of the author(s) and do not necessarily reflect the views of the NSF. 
\end{ack}

\bibliographystyle{plain}
\bibliography{%
biblio/barrier,%
biblio/books,%
biblio/adaptive,%
biblio/hybrid,%
biblio/mpc,%
biblio/nonlinear%
}

\appendix
\section{Supporting Constants}
\label{sec:app}
Given the compact set $\chi\subset\mathbb{R}^n$ and a continuous mapping $(\cdot)\,:\,\chi\rightarrow\mathbb{R}^N$, with $N\in\mathbb{N}$, define $\overline{\|(\cdot)\|}\coloneqq\sup_{x\in\chi}\|(\cdot)\|$. The constants used from \eqref{eq:suff_cond} are defined as { $\kappa \coloneqq \min\{\lambda_{\min}(Q),\tfrac{\underline{c}k_{c2}}{4},\tfrac{k_{a1}+k_{a2}}{4},\tfrac{K_{\theta}}{4}\}$, $\iota\coloneqq\tfrac{1}{2}\overline{\|\nabla V^*G_R\|}(\overline{\|\phi^T\nabla W + \nabla\varepsilon^T\|} + c_b\|\nabla B^T\|_{\infty}) + \tfrac{\iota_c^2}{2\underline{c}k_{c2}} + \tfrac{\iota_a^2}{2(k_{a1}+k_{a2})} + \tfrac{D_{\theta}^2}{2K_{\theta}}$, where $\iota_c\coloneqq \tfrac{k_{c1}3\sqrt{3}}{64\sqrt{\gamma_c}}\overline{\|G_R\|}\|\nabla B\|_{\infty}^2 + \tfrac{(k_{c1} + k_{c2})3\sqrt{3}}{16\sqrt{\gamma_c}}\overline{\|\Delta\|} + \tfrac{\iota_x}{\underline{\Gamma}}$, $\iota_{a}\coloneqq \tfrac{1}{2}\overline{\|\nabla V^* G_R\nabla\phi^T\|} +  k_{a2}\overline{\|W\|} + \tfrac{(k_{c1} + k_{c2})3\sqrt{3}}{64\sqrt{\gamma_c}}\overline{\|G_{\phi}\|}\overline{\|W\|}^2 + \iota_x$, $\iota_x\coloneqq \overline{\|\nabla W f\|} + \tfrac{1}{2}\overline{\|\nabla W G_R \nabla\phi^T W \|} + \tfrac{c_b}{2}\overline{\|\nabla W G_R\|}\|\nabla B^T\|_{\infty}$. The entries in the matrix $M$ from \eqref{eq:suff_cond} are defined as $\varphi_a\coloneqq \tfrac{(k_{c1} + k_{c2}3\sqrt{3})}{64\sqrt{\gamma_c}}\overline{\|G_\phi\|\|W\|} + \tfrac{1}{2}\overline{\|\nabla W G_R \nabla\phi^T\|}$, $\varphi_{ac}\coloneqq k_{a1} + \tfrac{(k_{c1} + k_{c2})3\sqrt{3}}{64\sqrt{\gamma_c}}\overline{\|G_{\phi}\| \|W\|} + \tfrac{c_bk_{c1}3\sqrt{3}}{32\sqrt{\gamma_c}}\overline{\|G_R\nabla\phi^T\|}\|\nabla B\|_{
\infty} + \tfrac{1}{2\underline{\Gamma}}\overline{\|\nabla W G_R\nabla\phi^T\|}$, $\varphi_{c\theta}\coloneqq \tfrac{(k_{c1} + k_{c2})3\sqrt{3}}{16\sqrt{\gamma_c}}\overline{\|W\nabla\phi Y\|}$.}

\end{document}